\pdfoutput=1
\documentclass[11pt]{article}

\usepackage[preprint]{acl}

\usepackage{times}
\usepackage{latexsym}
\usepackage[most]{tcolorbox}
\usepackage{xcolor}
\usepackage{booktabs}

\usepackage[T1]{fontenc}
\usepackage[utf8]{inputenc}

\usepackage{microtype}

\usepackage{inconsolata}

\usepackage{graphicx}
\usepackage[framemethod=default]{mdframed}
\newmdenv[
  linecolor=gray!60,
  linewidth=0.6pt,
  roundcorner=4pt,
  backgroundcolor=gray!5,
  skipabove=12pt,
  skipbelow=12pt,
  innertopmargin=6pt,
  innerbottommargin=6pt,
  innerleftmargin=8pt,
  innerrightmargin=8pt,
]{promptbox}
\title{Streamlining Industrial Contract Management with Retrieval-Augmented LLMs}

\author{Kristi Topollai\textsuperscript{1},
  Tolga Dimlioglu\textsuperscript{1}, Anna Choromanska\textsuperscript{1}, Simon Odie\textsuperscript{2}, Reginald Hui\textsuperscript{2}\\
  \textsuperscript{1}New York University\\  \textsuperscript{2}Consolidated
Edison Company of New York Inc. (Con Edison),\\
  \texttt{\{kt2664, td2249, ac5455\}@nyu.edu} \\
  \texttt{\{ODIES, huir\}@coned.com} }

\begin{document}
\maketitle
\begin{abstract}
Contract management involves reviewing and negotiating provisions, individual clauses that define rights, obligations, and terms of agreement. During this process, revisions to provisions are proposed and iteratively refined, some of which may be problematic or unacceptable. Automating this workflow is challenging due to the scarcity of labeled data and the abundance of unstructured legacy contracts. In this paper, we present a modular framework designed to streamline contract management through a retrieval-augmented generation (RAG) pipeline. Our system integrates synthetic data generation, semantic clause retrieval, acceptability classification, and reward‐based alignment to flag problematic revisions and generate improved alternatives. Developed and evaluated in collaboration with an industry partner, our system achieves over 80\% accuracy in both identifying and optimizing problematic revisions, demonstrating strong performance under real-world, low-resource conditions and offering a practical means of accelerating contract revision workflows.
\end{abstract}

\section{Introduction}

Contract management is a labor-intensive and critical process essential for risk management, operational efficiency, and regulatory compliance in both legal and business environments. Traditional approaches to contract analysis involve meticulously reviewing complex and lengthy documents to identify crucial clauses, obligations, exceptions, and potential risks. Such manual methods are not only time-consuming but also prone to human error, requiring significant legal expertise to accurately interpret nuanced language and cross-reference multiple documents to ensure compliance with relevant laws and standards. Furthermore, organizations frequently possess extensive legacy contractual data that remain largely unlabeled and unstructured, making them difficult to utilize  for informed decision-making.

\begin{figure}[t]
    \centering
    \includegraphics[width=\linewidth]{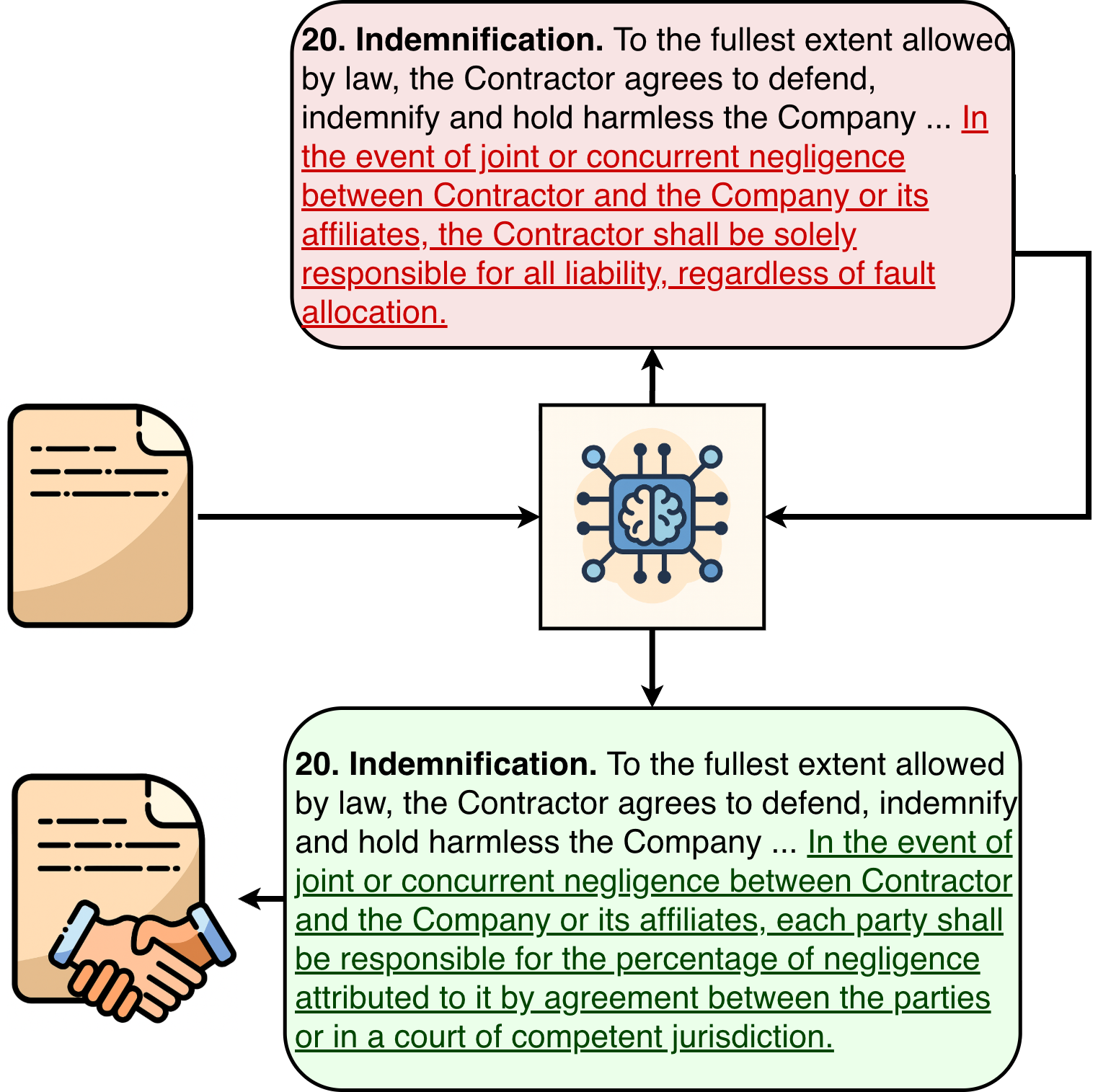}
    \caption{The system flags problematic clauses and rewrites them into acceptable revisions, reducing the risk of negotiation failure.}
    \vspace{-2em}
    \label{fig:overal_system}
\end{figure}

In this paper, we present a novel proof-of-concept framework for automating contract management and revision in settings with limited supervision and large volumes of legacy data. We evaluate the system on a real-world internal dataset provided by an industry partner in the utility sector. Our core contribution is an integrated pipeline that combines synthetic data generation, semantic clause retrieval, and acceptability classification to support systematic contract analysis. This framework enables the transformation of unstructured legacy contracts into structured, actionable insights, improving both the efficiency and reliability of identifying and addressing problematic revisions.

\begin{figure*}[t]
    \centering
    \includegraphics[width=\linewidth]{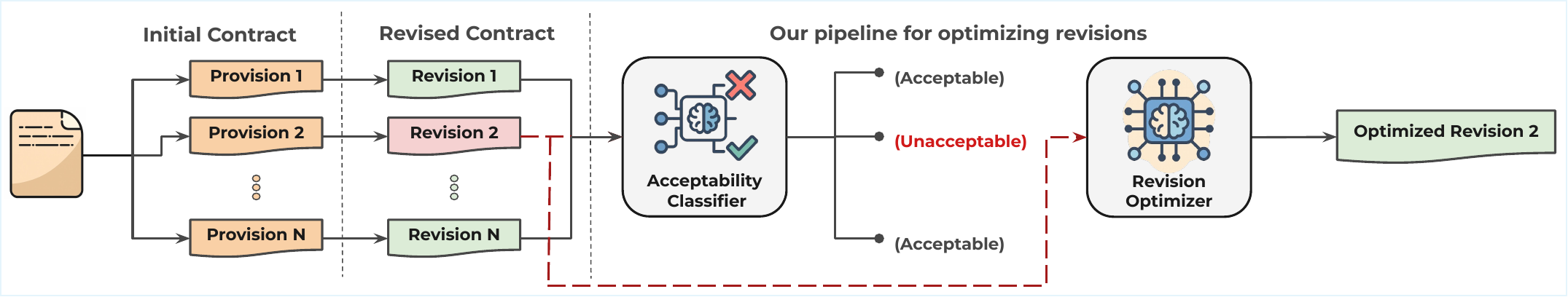}
    \caption{The structure of a contract. Our tool operates on each contract revision it identifies as problematic.}
    \label{fig:rag1}
\end{figure*}

At the core of our proposed solution is a Retrieval-Augmented Generation (RAG)~\cite{10.5555/3495724.3496517} methodology, designed to automatically identify unacceptable contract revisions, retrieve historically relevant clauses, and generate contextually appropriate amendments. By combining synthetic data creation, semantic-based retrieval, revision classification, and reinforcement-learning–based alignment of the generator to acceptability judgments, our approach reduces manual labor, mitigates human error, and enhances overall contractual consistency. This research demonstrates a practical path forward for organizations to effectively harness their extensive legacy contract data, transforming it into a dynamic resource for informed decision-making and streamlined contract management processes.

\section{Related Work}

Early efforts in automating contract analysis focused on rule-based and shallow learning methods for element extraction and clause classification. A benchmark of 3,500 contracts with 11 annotated element types (e.g., termination dates, parties, payments) was introduced in \cite{chalkidis2017extracting}, followed by work on obligation/prohibition extraction with hierarchical BiLSTMs \cite{chalkidis2018obligation} and unfair clause detection via deep models in the CLAUDETTE system \cite{lippi2019claudette}. Domain-specific resources include a lease-focused benchmark with ALeaseBERT \cite{leivaditi2020benchmark} and the large-scale CUAD dataset for clause extraction \cite{hendrycks2021cuad}.

Beyond extraction, several works tackle legal reasoning. A tax law dataset \cite{holzenberger2020dataset} showed symbolic solvers outperform neural models on entailment tasks. ContractNLI \cite{koreeda2021contractnli} frames clause-level review as document-level NLI, while LEGALBENCH \cite{guha2023legalbench} evaluates LLMs across 162 legal reasoning tasks. Recent work incorporates LLMs for generation and clause variation. \cite{lam2023applying} uses RAG to draft new clauses from negotiation intent, while \cite{narendra2024enhancing} applies Natural Language Inference (NLI) and RAG to detect deviations from contract templates and build clause libraries. In contrast, we focus on optimizing revisions to existing clauses during negotiation, enabling faster review and improved consistency. Our system combines synthetic data generation, semantic retrieval, and a lightweight acceptability classifier within a clause-level RAG pipeline. Unlike prior work limited to classification or extraction, our framework generates revised clauses likely to be accepted, operating effectively in low-resource settings with minimal supervision.

% I guess we need to mention more of ConEdison in this section
\section{Dataset}
Contracts are the backbone of modern business and legal relationships. They define how parties collaborate, what each side is responsible for, and what happens when things go wrong. These agreements are built from a set of provisions, or clauses, that outline specific terms such as scope of work, payment, liability, or termination. During negotiations, these provisions are often revised as parties, such as those involved in our industrial case study, seek to align the contract with their respective needs and constraints. Some revisions are acceptable and lead to a workable consensus, while others may introduce legal or operational risks and must be rejected. This process is time-consuming for both parties, and reaching an agreement can be one of the most resource-intensive parts of contract management.

\subsection{Internal Dataset}

Ideally, we would have access to the full lifecycle of contract formation, including all intermediate revisions to each provision and the rationale behind every change. However, we operate in a more constrained setting, with only limited supervision available. The labeled data used in our study was obtained internally and is provided in two forms.

First, we are provided with a small, curated set of fallback revisions: six acceptable and six unacceptable revisions for each of two provisions, totaling just 24 ground truth examples. Second, we leverage a collection of 20 previously negotiated contracts, 10 from the \textit{Standard Terms and Conditions for Service} and 10 from the \textit{Standard Terms and Conditions for Purchase of Equipment}. These documents contain tracked edits made during real-world negotiations. From these, we extract revised provisions and apply a weak labeling heuristic: edited provisions are treated as unacceptable (requiring correction), while non-edited provisions that differ from the original template are treated as acceptable. In total, our data has 287 acceptable and 143 unacceptable labeled revisions.

\subsection{Synthetic Data}

Given the limited size of our labeled dataset, relying solely on supervised learning is insufficient for training robust models. To overcome this, we leverage synthetic data generation with large language models (LLMs), a practical strategy for low-resource scenarios \cite{meng2022generating, Ye2022ZeroGenEZ, wang-etal-2023-self-instruct}. Prior work has shown that prompting LLMs to generate labeled examples can yield high-quality supervision for tasks like classification~\cite{li-etal-2023-synthetic}, NLI~\cite{hosseini-etal-2024-synthetic}, and summarization~\cite{chintagunta2021medically}, often rivaling human-labeled datasets in downstream performance. In our setting, synthetic clause revisions help augment the training set and improve generalization, particularly in learning patterns that distinguish acceptable from unacceptable revisions. This approach allows us to scale beyond the small set of manually curated fallback examples and tracked contract edits.

Due to data sensitivity constraints, all generation is performed locally, limiting the size of LLMs we can deploy. We experiment with models from the LLaMA 3 \cite{grattafiori2024llama} family, specifically the 8-billion parameter LLaMA 3.1 model and a AutoGPTQ~\cite{frantar2023gptq,li2025gptaq} int4-quantized variant  of the 70-billion parameter LLaMA 3.3 model. To generate synthetic labeled examples, we use a simple prompting strategy in which we provide the model demonstrations consisting of a contract provision followed by an acceptable and an unacceptable revision. An illustrative prompt can be found in the Appendix.

The generated set of synthetic revisions is further refined through a filtering process. We begin by encoding each synthetic revision using a general-purpose text embedding model \cite{li2023towards} and retrieving its nearest real revisions based on L2 distance in the embedding space. If the majority label of the retrieved neighbors disagrees with the synthetic label, the example is discarded. In the end we are left with around 27,000 synthetic revisions.

\begin{figure}[h]
    \centering
    \includegraphics[width=0.9\linewidth]{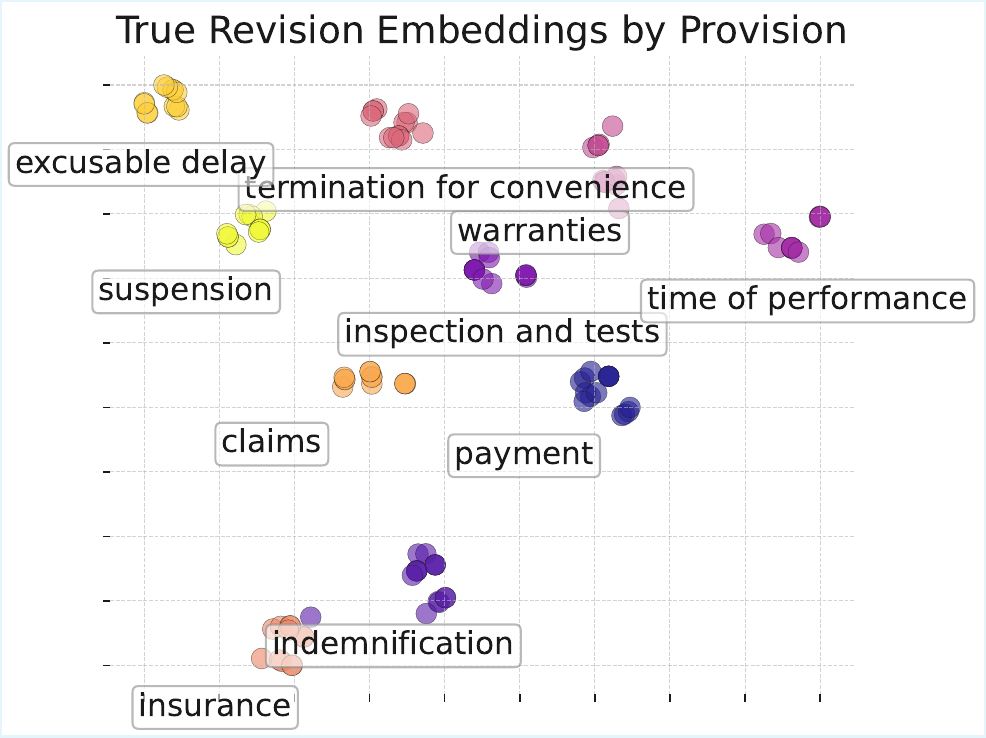}
    \caption{The revisions are clustered by their provision in the embedding space,}
    \label{fig:true_revisions}
\end{figure}

\begin{figure}[b]
    \centering
    \includegraphics[width=0.7\linewidth]{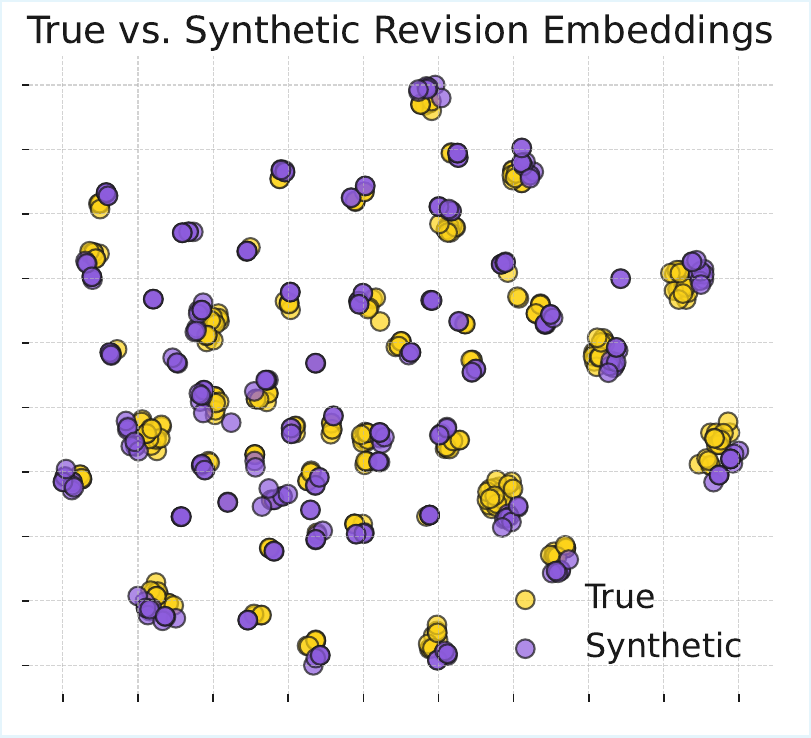}
    \caption{The t-SNE visualization demonstrates that real and synthetic revisions exhibit similar distributions in the embedding space.}
    \label{fig:embed_comparison}
\end{figure}
%A discussion of the different prompt formats we experimented with can be found in Section XXX, and details on filtering incorrect or inaccurate generations are provided in Section XXX of the supplementary material.

\begin{figure*}[t]
    \centering
    \includegraphics[width=\linewidth]{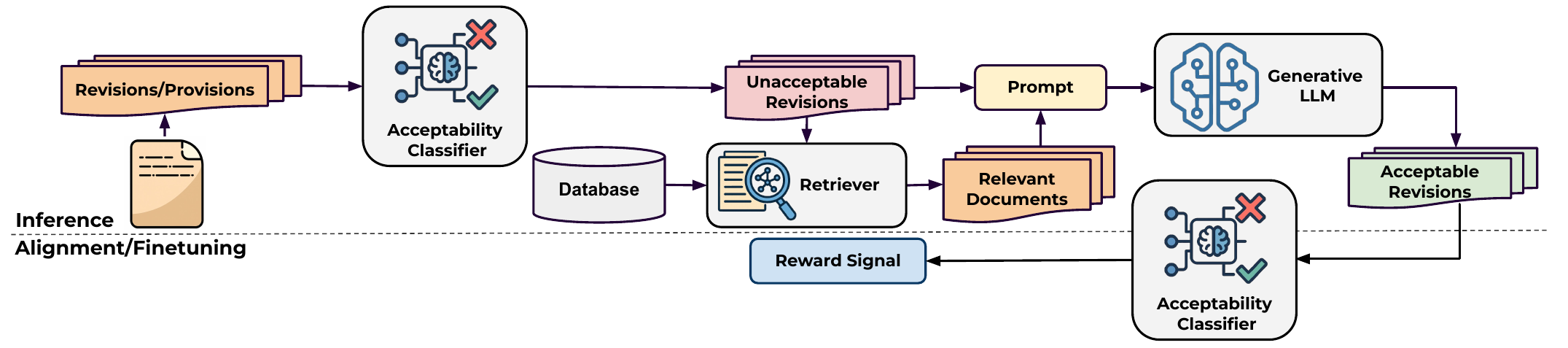}
    \caption{Our modular RAG-based pipeline. Using a frozen classifier for supervision allows end-to-end finetuning.}
    \label{fig:rag}
\end{figure*}

Figure~\ref{fig:embed_comparison} provides t-SNE visualization of the text embeddings and shows that synthetic embeddings preserve clause-type clustering, with a clear bimodal separation between acceptable and unacceptable revisions. Table~\ref{tab:datasets} supports our choice of the LLaMA 3.1 8B model over the int4-quantized 70B variant, as the former yields better Fréchet Inception Distance (FID) scores, adapted here to measure distributional similarity between real and synthetic text embeddings, under constrained resources. We also observe diminishing returns beyond three demonstrations per prompt, which we use in our final dataset to balance quality and prompt length.

\begin{table}[h]
\centering
\scriptsize
\begin{tabular}{@{}ll@{}}
\toprule
\textbf{Synthetic Dataset Generated by}         & \textbf{FID Score} \\ \midrule
Baseline (between two subsets of the real data) & 0.08               \\
INT4-Quantized LLaMA 3.3 70B (1 demonstration)  & 0.46               \\
LLaMA 3.1 8B (1 demonstration)                  & 0.31               \\
LLaMA 3.1 8B (3 demonstrations)                 & 0.14               \\
LLaMA 3.1 8B (5 demonstrations)                 & 0.16               \\ \bottomrule
\end{tabular}
\caption{Comparison between the synthetic datasets}
\label{tab:datasets}
\end{table}

\section{Method}
\subsection{Retrieval-Augmented Revision Optimization}

We propose a modular, (semi)-automatic system designed to assist legal professionals during contract negotiations. The tool accelerates the revision process by identifying potentially unacceptable clauses and generating improved alternatives that are more likely to reach consensus. To allow human oversight, our system is modular and designed to keep the legal expert in the loop while automating the most repetitive aspects of clause revision. The architecture consists of four key components: (1) a structured database of provisions and their historical revisions, (2) a semantic similarity retriever, (3) an acceptability classifier, and (4) a generative module. These modules are integrated into a retrieval-augmented generation (RAG) pipeline.

The pipeline begins by segmenting a contract into its constituent revised clauses. The acceptability classifier then flags revisions that are likely to be rejected. For each problematic clause, the retriever identifies similar revisions from the database and extracts other relevant provisions from the current contract to provide context. This set of retrieved documents is used to construct a prompt for the generative module, which rewrites the clause with the goal of increasing its likelihood of acceptance.

We leverage a pretrained LLaMA-3.1 8B model to generate optimized contract revisions under two complementary paradigms. First, in zero-shot inference mode, we simply chain the off-the-shelf retriever and generator without any additional training. Second, in reward-based alignment mode, we freeze both the retriever and our acceptability classifier, using the latter as a frozen reward model to refine only the generator via reinforcement learning. Concretely, we take the classifier’s positive-class probability as the reward signal and update the generator’s parameters with Proximal Policy Optimization (PPO) \cite{DBLP:journals/corr/SchulmanWDRK17}, a stable actor–critic algorithm tailored for large neural policies. This closely follows the Reinforcement Learning from Human Feedback (RLHF) paradigm \cite{christiano2017deep}, effectively aligning the generator’s outputs with the acceptability judgments encoded by our pretrained classifier. At inference time, the frozen retriever provides full-contract context, and the trained generator produces revisions that maximize the expected acceptability reward.

\begin{table*}[t]
\centering
\scriptsize
\begin{tabular}{@{}lllll@{}}
\toprule
\textbf{Retriever}                          & \textbf{Provision Retrieval} & \textbf{Top-10 Accuracy} & \textbf{Top-5 Accuracy} & \textbf{Top-1 Accuracy} \\ \midrule
legal-BERT                                  & 96.12\%                      & 61.24\%                  & 51.08\%                 & 29.63\%                 \\
legal-Longformer                            & 90.21\%                      & 46.50\%                  & 38.99\%                 & 22.45\%                 \\ \midrule
Qwen2-1.5B-Instruct                         & 99.90\%                      & 88.23\%                  & 80.34\%                 & 53.64\%                 \\
Qwen3-Embedding-4B                          & \textbf{99.94}\%                      & \textbf{90.17}\%                  & 84.01\%                 & 55.98\%                 \\ \midrule
+ Pretrained BGE-Reranker-Large             & \multicolumn{2}{l}{-}                                   & 52.22\%                 & 12.31\%                 \\
+ Finetuned BGE-Reranker-Large (1st method) & \multicolumn{2}{l}{-}                                   & 72.76\%                 & 51.19\%                 \\
+ Finetuned BGE-Reranker-Large (2nd method) & \multicolumn{2}{l}{-}                                   & \textbf{88.37}\%                 & \textbf{58.93}\%                 \\ \bottomrule
\end{tabular}
\caption{Retrieval accuracy between different embedding models and Qwen3 + Rerankers.}
\label{tab:retrieval}
\end{table*}

\subsection{Similarity Retriever}
\label{retriever}
The similarity retriever in our system serves two main purposes: (1) retrieving past contract revisions that resemble a given query, providing precedent for potential rewrites, and (2) identifying contextually related clauses within the same contract, which can influence how a revision is interpreted. In addition, due to the absence of manual similarity labels, we generate synthetic supervision by creating semantically equivalent paraphrases of each revision. This augmented dataset is used for both fine-tuning and evaluation.

\subsubsection{Retrieving Past Revisions}

Our database stores the text and embeddings of prior revisions. Given the document length, we experimented with embedding models with large context windows ($>2048$ tokens), and evaluated both general-purpose and legal-domain variants. Retrieval is performed via cosine similarity with the query revision, followed by reranking the top-$K$ hits using a cross-encoder.

While embedding model fine-tuning had minimal impact, due to limited data, fine-tuning the cross-encoder substantially improved results. We compare two strategies: \textbf{i)} binary classification (similar vs.\ not similar), and \textbf{ii)} graded similarity with soft labels: $y=1$ for paraphrases, $y=0.5$ for acceptable revisions of the same provision, $y=0.3$ for acceptable vs.\ unacceptable pairs of the same provision, and $y=0$ for unrelated provisions. The graded approach captures finer semantic distinctions critical for legal document retrieval.

\subsubsection{Intra-Contract Clause Retrieval}

To identify clauses within the same contract that are contextually related, we consider two approaches. The first leverages expert-provided labels for our internal contract templates. Alternatively, we adopt an automated strategy inspired by \citet{lam2023applying}. We begin by using our LLaMA model to extract keywords, key phrases, and explicit references to other clauses. Given the relatively small number of provisions per contract, we then apply a cross-encoder to all clause pairs and retain those whose similarity score exceeds a predefined threshold. This allows us to efficiently capture clause interdependencies, which are also important for generating context-aware revisions.

\subsection{Acceptability Classifier}
The acceptability classifier is a central component of our pipeline, responsible for flagging potentially problematic revisions. We frame this as a binary classification task, acceptable vs. unacceptable, and explore two complementary approaches.

The first leverages the zero-shot capabilities of a generative LLM (LLaMA in our case). We retrieve the top-$K$ semantically similar acceptable and unacceptable revisions (Section~\ref{retriever}) as demonstrations and prompt the model to classify the query revision. This approach offers interpretability through reasoning both in the input and output, facilitating collaboration with legal professionals.

The second approach employs a discriminative model trained on learned embeddings. Revisions are encoded using the same embedding model from the retrieval module and classified using logistic regression. To account for variation across provisions, we adopt an ensemble strategy: revisions are clustered into $K$ groups based on their embeddings, and a separate classifier is trained for each cluster. At inference time, the query is routed to its nearest cluster for prediction. The choice of $K$ balances generality and specialization.
 %The sensitivity analysis of the ensemble to the choice of $K$, as well as a comparison of the alternative acceptability classification methods we explored, can be found in Section XXX of the Appendix. 

\section{Experiments}
We first evaluate the retrieval and classification modules independently, detailing their architectures and evaluation protocols, before assessing the full pipeline's ability to generate more acceptable contract revisions.

\subsection{Similarity Retriever}
We evaluate the similarity retriever on our augmented synthetic revision dataset, where each original revision is paired with a semantically equivalent rephrased version. The evaluation metric is top-1 and top-K retrieval accuracy: the fraction of queries for which the correct rephrased version is retrieved among the top-1 or top-K results, respectively. We compare four embedding models:
\begin{itemize}
    \item \textbf{Legal-domain:} \textsc{Legal-BERT}~\cite{chalkidis-etal-2020-legal} and \textsc{Legal-Longformer}~\cite{mamakas-etal-2022-processing},
    \item \textbf{General-purpose models:} \textsc{Qwen2-1.5B-Instruct}~\cite{qwen2} and \textsc{Qwen3-Embedding-4B}~\cite{yang2025qwen3}, both of which are top-performing models on the MTEB benchmark~\cite{muennighoff-etal-2023-mteb}.
\end{itemize}

As shown in Table~\ref{tab:retrieval}, general-purpose models significantly outperform legal-domain models, primarily due to the latter’s limited context length, which hampers clause-level semantic understanding. Among the general-purpose models, Qwen3-Embedding-4B achieves the best top-1 and top-K retrieval accuracy and consistently retrieves semantically aligned revisions from the same provision. This evaluation is particularly challenging, as the synthetic dataset includes many near-duplicate revisions generated from a limited set of roughly 400 unique examples.

\paragraph{Reranking.}
We apply the \textsc{bge-reranker-large}~\cite{10.1145/3626772.3657878} cross-encoder to rerank the top-10 candidates retrieved by the best-performing embedding model, Qwen3-Embedding-4B. We evaluate both the off-the-shelf pretrained reranker and two versions finetuned using our proposed strategies. Surprisingly, as seen in Table~\ref{tab:retrieval}, the pretrained reranker fails to improve the initial ranking. Moreover, the binary classification-based finetuning strategy performs worse than using no reranker at all, leading to degraded top-1 accuracy. In contrast, the graded similarity-based finetuning substantially improves the reranking performance, outperforming both the baseline and the binary-trained variant.

\subsection{Acceptability Classifier}
\label{exp:classifier}
We evaluate the acceptability classifier using a train/test split on the synthetic dataset, with training performed exclusively on the synthetic training set. Evaluation is conducted on both the held-out synthetic test set and the original collection of 430 real revisions. For the embedding-based approach, we use the Qwen3-Embedding-4B model to generate embeddings, followed by a simple ensemble of logistic regression classifiers and we set the number of clusters $K=X$ through hyperparameter search.

\begin{table}[h]
\centering
\scriptsize
\begin{tabular}{@{}lllll@{}}
\toprule
\textbf{Classifier}            & \multicolumn{2}{l}{\textbf{Synthetic }} & \textbf{Original } &          \\ \midrule
                               & Train                & Test                 & Test                   & F1  \\ \midrule
Llama Zero-Shot (No clustering)               & \multicolumn{2}{l}{-}                       & 65.7\%                 & 0.682    \\
Logistic Ensemble (No clustering)  & 89.1\%               & 84.4\%               & 79.3\%                 & 0.851    \\
Logistic Ensemble (5 clusters) & 91.9\%               & 85.6\%               & 82.8\%                 & 0.878    \\
Logistic Ensemble (8 clusters) & 93.0\%               & 85.9\%               & \textbf{84.7\%}        & \textbf{0.889}    \\ \bottomrule
\end{tabular}
\caption{Comparison between acceptability classifiers.}
\label{tab:classifier}
\end{table}

The results can be found in Table~\ref{tab:classifier}. Interestingly, the zero-shot LLaMA-based approach struggles to reliably distinguish between acceptable and unacceptable revisions. In contrast, the embedding-based classifier performs significantly better. Moreover, its misclassifications tend to be associated with low-confidence (i.e., ambiguous) predictions, making it especially suitable for a semi-automatic setup, where uncertain cases can still be flagged for expert review.

\subsection{Retrieval-Augmented Revision Optimization}

Assessing the success of our revision optimization presents its own challenges. To automate evaluation at scale, we apply our frozen acceptability classifier to each post-optimization revision and report the fraction classified as acceptable. Although this metric depends on a classifier that is not perfect, it nevertheless provides a practical proxy when labeled data are scarce.

\begin{table}[h]
\centering
\scriptsize
\begin{tabular}{@{}ll@{}}
\toprule
\textbf{Method} & \shortstack{\textbf{Percentage of} \\ \textbf{Successful Optimizations}} \\
\midrule
\textbf{Zero-Shot Inference (Modular Pretraining)} &                                                                                                          \\
1 Demonstration                           & 59.1\%                                                                                                   \\
5 Demonstrations                          & 67.5\%                                                                                                   \\ \midrule
\textbf{Acceptability-based Alignment}             &                                                                                                          \\
1 Demonstration                           & 80.8\%                                                                                                   \\
5 Demonstrations                          & \textbf{81.9}\%                                                                                                   \\ \bottomrule

\end{tabular}

\caption{Comparison between our two RAG pipelines for revision optimization.}
\label{tab:rag}
\end{table}

Detailed results appear in Table \ref{tab:rag} and the Supplement. In zero-shot mode, performance improves as we increase the number of in-context demonstrations, peaking at four examples before plateauing. Under our acceptability-based alignment procedure, reinforcement-learning refinement further boosts the rate of acceptable revisions, bringing generated outputs closer to ground-truth acceptable edits. Although our reward model is trained on synthetic data and, these results highlight the promise of replacing it with expert-provided signals. %Incorporating legal-expert supervision into the reward function represents a natural next step toward even more robust, real-world alignment.

\section{Conclusion}
We presented a modular Retrieval-Augmented, LLM-based system for semi-automatic contract management, engineered to operate under minimal supervision and leverage large volumes of legacy contractual data. Our framework integrates synthetic data generation, semantic clause retrieval, and an ensemble-based acceptability classifier to detect and propose revisions for problematic contract clauses. Empirical results on internal industrial contracts demonstrate strong performance across semantic retrieval and revision optimization, showing that even highly domain-specific legal tasks can be meaningfully automated. Crucially, the pipeline is designed with a human-in-the-loop workflow that preserves legal rigor: automated tools offer candidate revisions, while domain experts make final decisions on edge cases or high-stakes clauses. This integration of automation and expert review improves productivity by focusing legal attention where it is most needed, without sacrificing contract integrity. Finally, the modular design allows for flexible customization across contracts, vendors, and legal regimes.

\clearpage
\newpage
\section*{Limitations}
While our system achieves strong overall performance, it currently struggles to capture fine-grained semantic changes within revision proposals, such as increases in contract budget or adjustments to milestone dates, that require contextual or numeric reasoning. Additionally, the pipeline does not incorporate the identity of the third-party vendor, which may influence whether a revision is acceptable. A clause modification that is acceptable to one vendor might be rejected by another, leading to potential mismatches in optimization. Nonetheless, the classifier reliably flags such vendor-sensitive revisions, ensuring they are routed to legal experts for targeted revision. This hybrid human-in-the-loop workflow still leads to significant productivity gains by automating the majority of routine clause adjustments. Future extensions could incorporate vendor identifiers or negotiation history into the dataset to further personalize optimization.


\bibliography{custom}

@inproceedings{chalkidis2017extracting,
  title={Extracting contract elements},
  author={Chalkidis, Ilias and Androutsopoulos, Ion and Michos, Achilleas},
  booktitle={Proceedings of the 16th edition of the International Conference on Artificial Intelligence and Law},
  pages={19--28},
  year={2017}
}

@article{chalkidis2018obligation,
  title={Obligation and prohibition extraction using hierarchical RNNs},
  author={Chalkidis, Ilias and Androutsopoulos, Ion and Michos, Achilleas},
  journal={arXiv preprint arXiv:1805.03871},
  year={2018}
}

@article{lippi2019claudette,
  title={CLAUDETTE: an automated detector of potentially unfair clauses in online terms of service},
  author={Lippi, Marco and Pa{\l}ka, Przemys{\l}aw and Contissa, Giuseppe and Lagioia, Francesca and Micklitz, Hans-Wolfgang and Sartor, Giovanni and Torroni, Paolo},
  journal={Artificial Intelligence and Law},
  volume={27},
  pages={117--139},
  year={2019},
  publisher={Springer}
}

@article{leivaditi2020benchmark,
  title={A benchmark for lease contract review},
  author={Leivaditi, Spyretta and Rossi, Julien and Kanoulas, Evangelos},
  journal={arXiv preprint arXiv:2010.10386},
  year={2020}
}

@article{holzenberger2020dataset,
  title={A dataset for statutory reasoning in tax law entailment and question answering},
  author={Holzenberger, Nils and Blair-Stanek, Andrew and Van Durme, Benjamin},
  journal={arXiv preprint arXiv:2005.05257},
  year={2020}
}

@article{hendrycks2021cuad,
  title={CUAD: an expert-annotated NLP dataset for legal contract review},
  author={Hendrycks, Dan and Burns, Collin and Chen, Anya and Ball, Spencer},
  journal={arXiv preprint arXiv:2103.06268},
  year={2021}
}

@article{koreeda2021contractnli,
  title={ContractNLI: A dataset for document-level natural language inference for contracts},
  author={Koreeda, Yuta and Manning, Christopher D},
  journal={arXiv preprint arXiv:2110.01799},
  year={2021}
}

@inproceedings{narendra2024enhancing,
  title={Enhancing Contract Negotiations with LLM-Based Legal Document Comparison},
  author={Narendra, Savinay and Shetty, Kaushal and Ratnaparkhi, Adwait},
  booktitle={Proceedings of the Natural Legal Language Processing Workshop 2024},
  pages={143--153},
  year={2024}
}

@inproceedings{lam2023applying,
  title={Applying Large Language Models for Enhancing Contract Drafting},
  author={Lam, Kwok-Yan and Cheng, Victor CW and Yeong, Zee Kin},
  booktitle={LegalAIIA@ ICAIL},
  pages={70--80},
  year={2023}
}

@article{guha2023legalbench,
  title={Legalbench: A collaboratively built benchmark for measuring legal reasoning in large language models},
  author={Guha, Neel and Nyarko, Julian and Ho, Daniel and R{\'e}, Christopher and Chilton, Adam and Chohlas-Wood, Alex and Peters, Austin and Waldon, Brandon and Rockmore, Daniel and Zambrano, Diego and others},
  journal={Advances in Neural Information Processing Systems},
  volume={36},
  pages={44123--44279},
  year={2023}
}

@article{meng2022generating,
  title={Generating training data with language models: Towards zero-shot language understanding},
  author={Meng, Yu and Huang, Jiaxin and Zhang, Yu and Han, Jiawei},
  journal={Advances in Neural Information Processing Systems},
  volume={35},
  pages={462--477},
  year={2022}
}

@inproceedings{wang-etal-2023-self-instruct,
    title = "Self-Instruct: Aligning Language Models with Self-Generated Instructions",
    author = "Wang, Yizhong  and
      Kordi, Yeganeh  and
      Mishra, Swaroop  and
      Liu, Alisa  and
      Smith, Noah A.  and
      Khashabi, Daniel  and
      Hajishirzi, Hannaneh",
    editor = "Rogers, Anna  and
      Boyd-Graber, Jordan  and
      Okazaki, Naoaki",
    booktitle = "Proceedings of the 61st Annual Meeting of the Association for Computational Linguistics (Volume 1: Long Papers)",
    month = jul,
    year = "2023",
    address = "Toronto, Canada",
    publisher = "Association for Computational Linguistics",
    url = "https://aclanthology.org/2023.acl-long.754/",
    doi = "10.18653/v1/2023.acl-long.754",
    pages = "13484--13508"
}

@inproceedings{Ye2022ZeroGenEZ,
  title={ZeroGen: Efficient Zero-shot Learning via Dataset Generation},
  author={Jiacheng Ye and Jiahui Gao and Qintong Li and Hang Xu and Jiangtao Feng and Zhiyong Wu and Tao Yu and Lingpeng Kong},
  booktitle={Conference on Empirical Methods in Natural Language Processing},
  year={2022},
  url={https://api.semanticscholar.org/CorpusID:246867045}
}

@article{li2023towards,
  title={Towards general text embeddings with multi-stage contrastive learning},
  author={Li, Zehan and Zhang, Xin and Zhang, Yanzhao and Long, Dingkun and Xie, Pengjun and Zhang, Meishan},
  journal={arXiv preprint arXiv:2308.03281},
  year={2023}
}

@inproceedings{10.5555/3495724.3496517,
author = {Lewis, Patrick and Perez, Ethan and Piktus, Aleksandra and Petroni, Fabio and Karpukhin, Vladimir and Goyal, Naman and K\"{u}ttler, Heinrich and Lewis, Mike and Yih, Wen-tau and Rockt\"{a}schel, Tim and Riedel, Sebastian and Kiela, Douwe},
title = {Retrieval-augmented generation for knowledge-intensive NLP tasks},
year = {2020},
isbn = {9781713829546},
publisher = {Curran Associates Inc.},
address = {Red Hook, NY, USA},
booktitle = {Proceedings of the 34th International Conference on Neural Information Processing Systems},
articleno = {793},
numpages = {16},
location = {Vancouver, BC, Canada},
series = {NIPS '20}
}

@article{grattafiori2024llama,
  title={The llama 3 herd of models},
  author={Grattafiori, Aaron and Dubey, Abhimanyu and Jauhri, Abhinav and Pandey, Abhinav and Kadian, Abhishek and Al-Dahle, Ahmad and Letman, Aiesha and Mathur, Akhil and Schelten, Alan and Vaughan, Alex and others},
  journal={arXiv preprint arXiv:2407.21783},
  year={2024}
}

@inproceedings{chalkidis-etal-2020-legal,
    title = "{LEGAL}-{BERT}: The Muppets straight out of Law School",
    author = "Chalkidis, Ilias  and
      Fergadiotis, Manos  and
      Malakasiotis, Prodromos  and
      Aletras, Nikolaos  and
      Androutsopoulos, Ion",
    editor = "Cohn, Trevor  and
      He, Yulan  and
      Liu, Yang",
    booktitle = "Findings of the Association for Computational Linguistics: EMNLP 2020",
    month = nov,
    year = "2020",
    address = "Online",
    publisher = "Association for Computational Linguistics",
    url = "https://aclanthology.org/2020.findings-emnlp.261/",
    doi = "10.18653/v1/2020.findings-emnlp.261",
    pages = "2898--2904"
}

@inproceedings{mamakas-etal-2022-processing,
    title = "Processing Long Legal Documents with Pre-trained Transformers: Modding {L}egal{BERT} and Longformer",
    author = "Mamakas, Dimitris  and
      Tsotsi, Petros  and
      Androutsopoulos, Ion  and
      Chalkidis, Ilias",
    editor = "Aletras, Nikolaos  and
      Chalkidis, Ilias  and
      Barrett, Leslie  and
      Goanț{\u{a}}, C{\u{a}}t{\u{a}}lina  and
      Preoțiuc-Pietro, Daniel",
    booktitle = "Proceedings of the Natural Legal Language Processing Workshop 2022",
    month = dec,
    year = "2022",
    address = "Abu Dhabi, United Arab Emirates (Hybrid)",
    publisher = "Association for Computational Linguistics",
    url = "https://aclanthology.org/2022.nllp-1.11/",
    doi = "10.18653/v1/2022.nllp-1.11",
    pages = "130--142"
}

@article{qwen2,
      title={Qwen2 Technical Report}, 
      author={An Yang and Baosong Yang and Binyuan Hui and Bo Zheng and Bowen Yu and Chang Zhou and Chengpeng Li and Chengyuan Li and Dayiheng Liu and Fei Huang and Guanting Dong and Haoran Wei and Huan Lin and Jialong Tang and Jialin Wang and Jian Yang and Jianhong Tu and Jianwei Zhang and Jianxin Ma and Jin Xu and Jingren Zhou and Jinze Bai and Jinzheng He and Junyang Lin and Kai Dang and Keming Lu and Keqin Chen and Kexin Yang and Mei Li and Mingfeng Xue and Na Ni and Pei Zhang and Peng Wang and Ru Peng and Rui Men and Ruize Gao and Runji Lin and Shijie Wang and Shuai Bai and Sinan Tan and Tianhang Zhu and Tianhao Li and Tianyu Liu and Wenbin Ge and Xiaodong Deng and Xiaohuan Zhou and Xingzhang Ren and Xinyu Zhang and Xipin Wei and Xuancheng Ren and Yang Fan and Yang Yao and Yichang Zhang and Yu Wan and Yunfei Chu and Yuqiong Liu and Zeyu Cui and Zhenru Zhang and Zhihao Fan},
      journal={arXiv preprint arXiv:2407.10671},
      year={2024}
}

@article{yang2025qwen3,
  title={Qwen3 technical report},
  author={Yang, An and Li, Anfeng and Yang, Baosong and Zhang, Beichen and Hui, Binyuan and Zheng, Bo and Yu, Bowen and Gao, Chang and Huang, Chengen and Lv, Chenxu and others},
  journal={arXiv preprint arXiv:2505.09388},
  year={2025}
}

@inproceedings{muennighoff-etal-2023-mteb,
    title = "{MTEB}: Massive Text Embedding Benchmark",
    author = "Muennighoff, Niklas  and
      Tazi, Nouamane  and
      Magne, Loic  and
      Reimers, Nils",
    editor = "Vlachos, Andreas  and
      Augenstein, Isabelle",
    booktitle = "Proceedings of the 17th Conference of the European Chapter of the Association for Computational Linguistics",
    month = may,
    year = "2023",
    address = "Dubrovnik, Croatia",
    publisher = "Association for Computational Linguistics",
    url = "https://aclanthology.org/2023.eacl-main.148/",
    doi = "10.18653/v1/2023.eacl-main.148",
    pages = "2014--2037"
}

@inproceedings{10.1145/3626772.3657878,
author = {Xiao, Shitao and Liu, Zheng and Zhang, Peitian and Muennighoff, Niklas and Lian, Defu and Nie, Jian-Yun},
title = {C-Pack: Packed Resources For General Chinese Embeddings},
year = {2024},
isbn = {9798400704314},
publisher = {Association for Computing Machinery},
address = {New York, NY, USA},
url = {https://doi.org/10.1145/3626772.3657878},
doi = {10.1145/3626772.3657878},
booktitle = {Proceedings of the 47th International ACM SIGIR Conference on Research and Development in Information Retrieval},
pages = {641–649},
numpages = {9},
location = {Washington DC, USA},
series = {SIGIR '24}
}

@inproceedings{li-etal-2023-synthetic,
    title = "Synthetic Data Generation with Large Language Models for Text Classification: Potential and Limitations",
    author = "Li, Zhuoyan  and
      Zhu, Hangxiao  and
      Lu, Zhuoran  and
      Yin, Ming",
    editor = "Bouamor, Houda  and
      Pino, Juan  and
      Bali, Kalika",
    booktitle = "Proceedings of the 2023 Conference on Empirical Methods in Natural Language Processing",
    month = dec,
    year = "2023",
    address = "Singapore",
    publisher = "Association for Computational Linguistics",
    url = "https://aclanthology.org/2023.emnlp-main.647/",
    doi = "10.18653/v1/2023.emnlp-main.647",
    pages = "10443--10461"
}

@inproceedings{hosseini-etal-2024-synthetic,
    title = "A synthetic data approach for domain generalization of {NLI} models",
    author = "Hosseini, Mohammad Javad  and
      Petrov, Andrey  and
      Fabrikant, Alex  and
      Louis, Annie",
    editor = "Ku, Lun-Wei  and
      Martins, Andre  and
      Srikumar, Vivek",
    booktitle = "Proceedings of the 62nd Annual Meeting of the Association for Computational Linguistics (Volume 1: Long Papers)",
    month = aug,
    year = "2024",
    address = "Bangkok, Thailand",
    publisher = "Association for Computational Linguistics",
    url = "https://aclanthology.org/2024.acl-long.120/",
    doi = "10.18653/v1/2024.acl-long.120",
    pages = "2212--2226"
}

@inproceedings{chintagunta2021medically,
  title={Medically aware GPT-3 as a data generator for medical dialogue summarization},
  author={Chintagunta, Bharath and Katariya, Namit and Amatriain, Xavier and Kannan, Anitha},
  booktitle={Machine Learning for Healthcare Conference},
  pages={354--372},
  year={2021},
  organization={PMLR}
}

@inproceedings{frantar2023gptq,
  title = {GPTQ: Accurate Post-Training Quantization for Generative Pre-trained Transformers},
  author = {Frantar, Elias and Ashkboos, Saleh and Hoefler, Torsten and Alistarh, Dan},
  booktitle = {Proceedings of the 11th International Conference on Learning Representations (ICLR)},
  year = {2023},
  url = {https://openreview.net/forum?id=-N2l2qTfyy},
  note = {Published at ICLR 2023}
}

@inproceedings{li2025gptaq,
  title = {GPTAQ: Efficient Finetuning-Free Quantization for Asymmetric Calibration},
  author = {Li, Yuhang and Yin, Ruokai and Lee, Donghyun and Xiao, Shiting and Panda, Priyadarshini},
  booktitle = {Proceedings of the 42nd International Conference on Machine Learning (ICML)},
  pages = {},
  year = {2025},
  organization = {PMLR},
  volume = {267},
  address = {Vancouver, Canada},
  copyright = {2025},
  note = {Finetuning-Free Quantization for Asymmetric Calibration}
}

@article{DBLP:journals/corr/SchulmanWDRK17,
  author       = {John Schulman and
                  Filip Wolski and
                  Prafulla Dhariwal and
                  Alec Radford and
                  Oleg Klimov},
  title        = {Proximal Policy Optimization Algorithms},
  journal      = {CoRR},
  volume       = {abs/1707.06347},
  year         = {2017},
  url          = {http://arxiv.org/abs/1707.06347},
  eprinttype    = {arXiv},
  eprint       = {1707.06347},
  bibsource    = {dblp computer science bibliography, https://dblp.org}
}

@article{christiano2017deep,
  title={Deep reinforcement learning from human preferences},
  author={Christiano, Paul F and Leike, Jan and Brown, Tom and Martic, Miljan and Legg, Shane and Amodei, Dario},
  journal={Advances in neural information processing systems},
  volume={30},
  year={2017}
}

\newpage
\appendix
\label{sec:appendix}

\section{Hardware and Environment}
\label{sec:hardware}
Due to the sensitivity of the internal dataset, all inference and fine-tuning were performed locally on private machines. Specifically, our experiments were conducted using two NVIDIA RTX 3090 GPUs, providing a total of 48GB of VRAM. This setup allowed us to run all LLMs used in our pipeline with full FP16 precision.

\section{Prompts}

\begin{promptbox}
\footnotesize
\textbf{Synthetic Revisions Prompt} \\
Use the following pairs of provisions and fallback revisions to understand what constitutes an acceptable and unacceptable revision. Then provide revisions for the given query provision.\\

\textbf{Demonstration 1} \\
\textbf{Provision:} [Random provision text] \\
\textbf{Acceptable revision:} [Acceptable revision ] \\
\textbf{Unacceptable revision:} [Unacceptable revision] \\[0.8em]
$\vdots$ \\[0.5em]

\textbf{Demonstration N} \\
\textbf{Provision:} [Random provision text] \\
\textbf{Acceptable revision:} [Acceptable revision] \\
\textbf{Unacceptable revision:} [Unacceptable revision] \\[0.5em]
\textbf{Query Provision:} [Query provision text]
\end{promptbox}

\vspace{-1.5em}

\begin{promptbox}
\footnotesize
\textbf{Revision Optimization Prompt} \\
Use the following examples of provisions and their revisions to learn how to transform unacceptable revisions into acceptable ones. Then, provide revised versions for the given query unacceptable revision. You are also provided with clauses from the same contract that may be contextually relevant to the query. Incorporate their meaning and constraints when rewriting. \\

\textbf{Demonstration 1} \\
\textbf{Provision:} [Provision text A] \\
\textbf{Unacceptable revision:} [Revised text A] \\
\textbf{Acceptable revision:} [Corrected version of A] \\

\textbf{Demonstration 2} \\
\textbf{Provision:} [Provision text B] \\
\textbf{Unacceptable revision:} [Revised text B] \\
\textbf{Acceptable revision:} [Corrected version of B] \\

$\vdots$ \\[0.5em]

\textbf{Related Clauses (from current contract):} \\
\textbf{Related clause:} [Related clause 1] \\
\textbf{Related clause:}  [Related clause 1]\\
$\vdots$

\textbf{Query Unacceptable Revision:} [Unacceptable Revision text Q] \\
\textbf{Optimized Unacceptable Version:} \\
\end{promptbox}

\begin{promptbox}
\footnotesize
\textbf{Rephrasing Prompt for Semantic Equivalence} \\
Rephrase the following contract clause revision so that it is semantically identical but expressed using different wording. Do not change the meaning, intent, or legal interpretation of the revision. Ensure the rephrasing retains the same level of formality and contractual tone.\\

\textbf{Original Revision:} [Original revision text]\\[0.5em]
\textbf{Rephrased Revision:} 
\end{promptbox}
\vspace{-1.5em}
\begin{promptbox}
\footnotesize
\textbf{Clause Dependency Extraction Prompt} \\
Given the contract text below, analyze the specified clause to extract: \\
(1) The key terms and phrases that summarize its content. \\
(2) Any explicit or implicit references to other clauses within the same contract (e.g. “as described in Section 5”, “subject to Clause 10”). \\
Return the output in JSON format with the keys \texttt{"keywords"}, \texttt{"key\_phrases"}, and \texttt{"references"}. Do not modify the text of the clause. \\

\textbf{Full Contract:} 
$[$Insert full or partial contract text here$]$\\[0.5em]

\textbf{Target Clause:} 
$[$Insert clause to analyze here$]$\\[0.5em]

\textbf{Output:}
\begin{verbatim}
{
  "keywords": [...],
  "key_phrases": [...],
  "references": [...]
}
\end{verbatim}
\end{promptbox}
\vspace{-1.5em}
\begin{promptbox}
\footnotesize
\textbf{Zero-Shot Acceptability Classification Prompt} \\
Below are examples of contract clause revisions labeled as either acceptable or unacceptable. Analyze the patterns in these examples and determine whether the given query revision should be classified as \texttt{ACCEPTABLE} or \texttt{UNACCEPTABLE}. Provide a brief justification for your classification. \\

\textbf{Demonstration 1} \\
\textbf{Revision:} [Revised text A] \\
\textbf{Label:} ACCEPTABLE \\[0.3em]

\textbf{Demonstration 2} \\
\textbf{Revision:} [Revised text B] \\
\textbf{Label:} UNACCEPTABLE \\[0.3em]

$\vdots$ \\[0.5em]

\textbf{Demonstration N} \\
\textbf{Revision:} [Revised text N] \\
\textbf{Label:} ACCEPTABLE \\[0.8em]

\textbf{Query Revision:} [Revision to classify]\\[0.5em]

\textbf{Output:}
\begin{verbatim}
Label: ACCEPTABLE
Justification: [Explain the decision]
\end{verbatim}
\end{promptbox}

\section{Original Dataset}
\begin{figure}[h]
\centering
\includegraphics[width=0.75\linewidth]{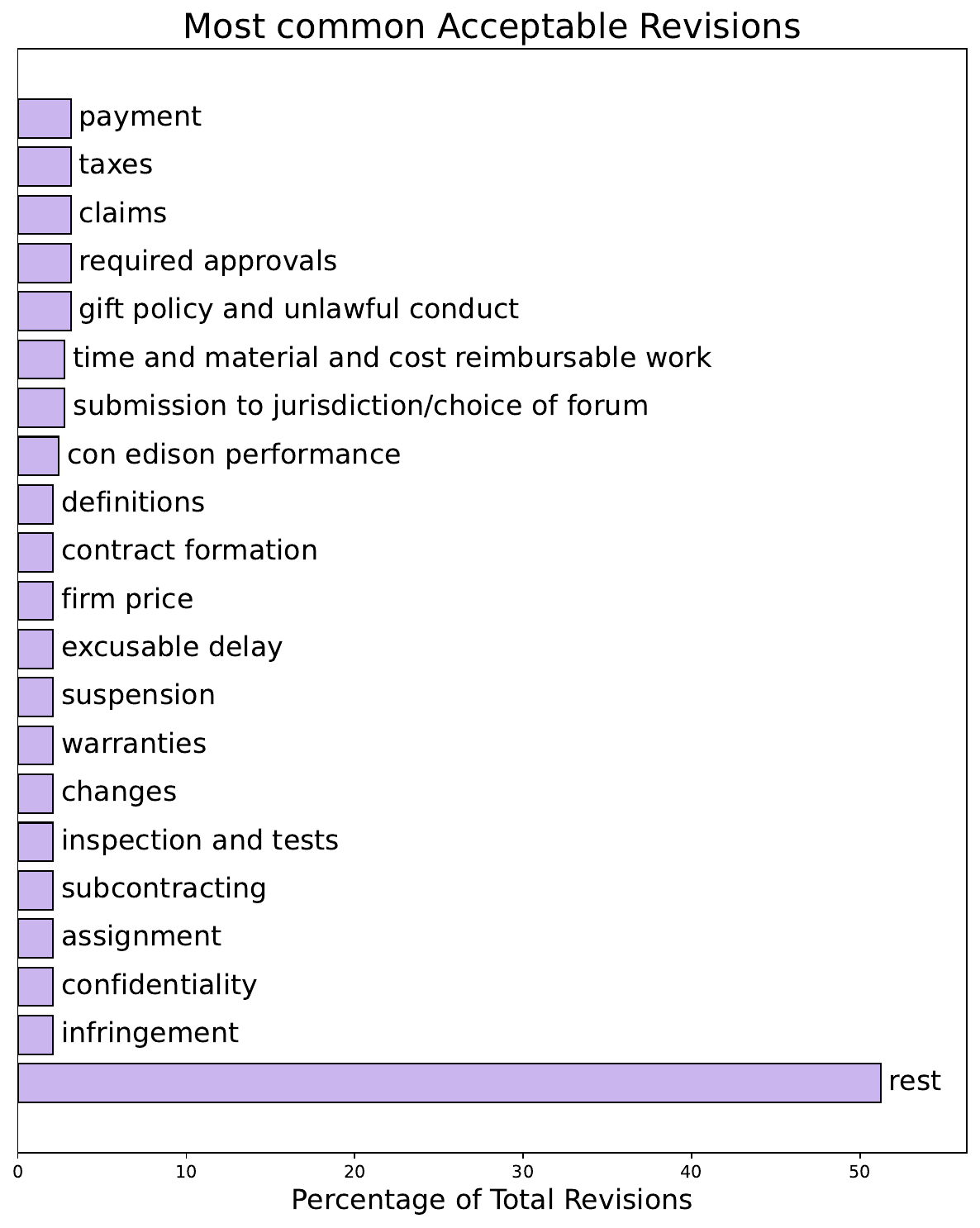}
\caption{Distribution of acceptable revisions across provisions. Each bar indicates the percentage of acceptable revisions contributed by that provision.}
\label{fig:acceptables}
\end{figure}

\begin{figure}[h]
\centering
\includegraphics[width=0.75\linewidth]{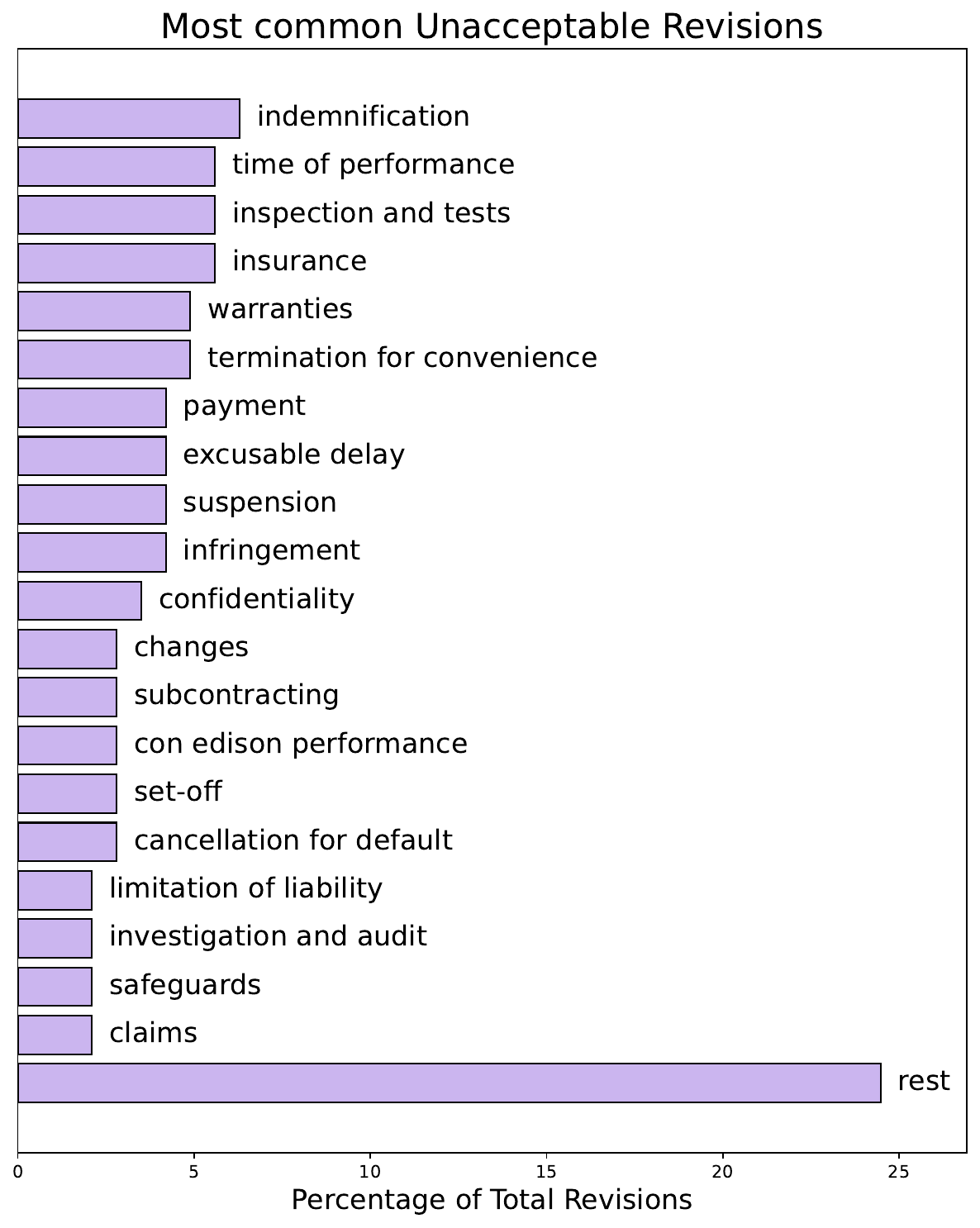}
\caption{Distribution of unacceptable revisions across provisions. A small subset of provisions accounts for the majority of unacceptable cases.}
\label{fig:unacceptables}
\end{figure}

Among all provisions, only a subset is frequently revised, and within this subset, certain provisions are disproportionately associated with unacceptable revisions. In fact, 75\% of unacceptable revisions fall under just a handful of provisions. Figures~\ref{fig:acceptables} and~\ref{fig:unacceptables} visualize the distribution of acceptable and unacceptable revisions, respectively, highlighting that while acceptable revisions are relatively evenly distributed, unacceptable revisions are concentrated in specific provision types such as \textit{indemnification}, \textit{time of performance}, and \textit{insurance}.

\vspace{-2em}

\section{Hyperparameters and Training Settings}

This section summarizes the key hyperparameters and configuration details used across different components of our system for reproducibility.

\begin{table}[h]
\centering
\footnotesize
\begin{tabular}{l|l}
\toprule
\textbf{Parameter} & \textbf{Value} \\
\midrule
Model & LLaMA 3.1–8B Instruct \\
Max new tokens & 8192\\
Temperature & 0.8\\
Top-k & 50\\
Top-p & 0.9\\
Number of demonstrations ($K$) & 1/2/\textbf{3}/4/5/7/9\\
Embedding model for filtering & Qwen3-Embedding-4B \\
Neighbors used for filtering ($k$) & 10/\textbf{20}/40\\
\bottomrule
\end{tabular}
\caption{Hyperparameters for synthetic revision generation.}
\end{table}

\begin{table}[h]
\centering
\footnotesize

\begin{tabular}{l|l}
\toprule
\textbf{Component} & \textbf{Setting} \\
\midrule
Embedding model & Qwen3-Embedding-4B \\
Similarity metric & \textbf{Cosine}/ L2 \\
Top-$K$ candidates retrieved & 5/\textbf{10}/20\\
\midrule
Reranker model & BGE-Reranker-Large \\
Finetuning objective & \textbf{Multiclass}/Binary Classification\\
Training epochs & 10\\
Optimizer & AdamW\\
betas & (0.9, 0.999)\\
weight decaye & 0.1 \\
Batch size & 128 \\
Learning rate & 0.001\\
Similarity label scheme & \textbf{\{1.0, 0.5, 0.3, 0.0\}}, \{1.0, 0.0\} \\
\bottomrule
\end{tabular}
\caption{Hyperparameters for retriever and reranker finetuning.}
\end{table}

\begin{table}[htp!]
\centering
\footnotesize
\begin{tabular}{l|l}
\toprule
\textbf{Component} & \textbf{Setting} \\
\midrule
Embedding model & Qwen3-Embedding-4B \\
Clustering algorithm & k-means \\
Number of clusters ($K$) & 3,5,\textbf{8},11\\
Classifier type & Logistic Regression \\
Train/val split ratio & 90/10\\
Routing metric & \textbf{Cosine}, L2 \\
\bottomrule
\end{tabular}
\caption{Hyperparameters for the acceptability classifier.}
\end{table}

\begin{table}[htp!]
\centering
\footnotesize
\begin{tabular}{l|l}
\toprule
\textbf{Parameter} & \textbf{Value} \\
\midrule
Retriever top-$K$ (past revisions) & 1/2/3/4/\textbf{5}/6/7\\
Max tokens for merged prompt & 128k\\
Temperature & 0.8\\
Top-k & 50\\
Top-p & 0.9\\
Max new tokens & 8192\\
\bottomrule
\end{tabular}
\caption{Hyperparameters for inference in the RAG pipeline.}
\end{table}

\begin{center}
\footnotesize
\begin{tabular}{l|l}
\toprule
\textbf{Parameter} & \textbf{Value} \\
\midrule
Policy model & LLaMA 3.1--8B Instruct \\
\hline
LoRA rank ($r$) & 8 \\
LoRA $\alpha$ & 32 \\
LoRA dropout & 0.05 \\
\hline
Reward model & Frozen acceptability classifier \\
Reward normalization & None \\
Batch size & 4 \\
PPO epochs & 4 \\
Learning rate & $1\!\times\!10^{-5}$ \\
Discount factor ($\gamma$) & 1.0 \\
Clip parameter ($\epsilon$) & 0.2 \\
Entropy coefficient & 0.01 \\
KL penalty coefficient & 0.1 \\
Max sequence length & 8192 tokens \\
Gradient clipping & 1.0 (global norm) \\
\bottomrule
\end{tabular}
\captionof{table}{Hyperparameters for PPO.}
\end{center}

%\newpage
%\clearpage
%\section{Retriever}

%\subsection{Paraphrased Dataset}

%As mentioned in the main text, to evaluate the retriever and to finetune the cross-encoder, we make use of a second synthetic dataset, which consists of paraphrased but semantically identical versions of each revision in the synthetic dataset. To paraphrase we again use the same LLaMA 3.1 8B

\end{document}